\documentclass[letterpaper]{article} 
\usepackage{aaai25}  
\usepackage{times}  
\usepackage{helvet}  
\usepackage{courier}  
\usepackage[hyphens]{url}  
\usepackage{graphicx} 
\usepackage{natbib}  
\usepackage{caption} 
\usepackage{algorithm}

\usepackage{newfloat}
\usepackage{listings}

\usepackage{algorithm}
\usepackage{algpseudocode}

\usepackage{times}
\usepackage{latexsym}
\usepackage{amsmath}
\usepackage{subcaption}
\usepackage{float}
\usepackage{algorithm}
\usepackage{algpseudocode}
\usepackage{booktabs}
\usepackage{float}
\usepackage{graphicx}
\usepackage[utf8]{inputenc}
\usepackage{microtype}
\usepackage{booktabs}
\usepackage{multirow}
\usepackage{array}
\usepackage[tableposition=top]{caption}
\usepackage{inconsolata}
\usepackage{amsfonts}
\usepackage{graphicx}
\usepackage{graphicx}
\usepackage{booktabs}
\usepackage{multirow}
\usepackage{adjustbox}
\usepackage{enumitem}

\DeclareCaptionStyle{ruled}{labelfont=normalfont,labelsep=colon,strut=off} 
\floatstyle{ruled}
\newfloat{listing}{tb}{lst}{}
\floatname{listing}{Listing}
\nocopyright 

\title{PagedEviction: 
Structured Block-wise KV Cache Pruning \\ for Efficient Large Language Model Inference}

\author{
  Krishna Teja Chitty-Venkata\equalcontrib\textsuperscript{\rm 1},
  Jie Ye\equalcontrib\textsuperscript{\rm 2},
  Xian-He Sun\textsuperscript{\rm 2},
  Anthony Kougkas\textsuperscript{\rm 2},\\
  Murali Emani\textsuperscript{\rm 1},
  Venkatram Vishwanath\textsuperscript{\rm 1},
  Bogdan Nicolae\textsuperscript{\rm 1}
}
\affiliations{
  \textsuperscript{\rm 1}Argonne National Laboratory, Lemont, IL, USA
  \textsuperscript{\rm 2}Illinois Institute of Technology, Chicago, IL, USA \\
  \{krishnat, memani, venkat, bnicolae\}@anl.gov, jye20@hawk.iit.edu, \{sun, akougkas\}@iit.edu 
}

\usepackage{bibentry}

\begin{document}

\maketitle

\begin{abstract}


KV caching significantly improves the efficiency of Large Language Model (LLM) inference by storing attention states from previously processed tokens, enabling faster generation of subsequent tokens. However, as sequence length increases, the KV cache quickly becomes a major memory bottleneck. To address this, we propose PagedEviction, a novel fine-grained, structured KV cache pruning strategy that enhances the memory efficiency of vLLM’s PagedAttention. Unlike existing approaches that rely on attention-based token importance or evict tokens across different vLLM pages, PagedEviction introduces an efficient block-wise eviction algorithm tailored for paged memory layouts. Our method integrates seamlessly with PagedAttention without requiring any modifications to its CUDA attention kernels. We evaluate PagedEviction across Llama-3.1-8B-Instruct, Llama-3.2-1B-Instruct, and Llama-3.2-3B-Instruct models on the LongBench benchmark suite, demonstrating improved memory usage with better accuracy than baselines on long context tasks. 

\end{abstract}

\section{Introduction}
\label{sec:intro}

Large Language Models (LLMs) have revolutionized the field of Natural Language Processing and are capable of understanding and generating human-like text. These models, trained on vast amounts of data, are capable of performing a wide range of language and summarizing tasks. The state-of-the-art (SOTA) LLMs are exploding to large sizes, include GPT \cite{brown2020language}, LLaMA \cite{touvron2023llama}, and DeepSeek \cite{dai2024deepseekmoe, guo2025deepseek}. \looseness=-1 

LLM inference is an iterative process that involves incremental computations, reusing a significant portion of previous attention results, which are stored in the form of KV Cache, thereby accelerating autoregressive generation. However, this comes at a high memory cost: the KV cache grows linearly with sequence length and can even rival or exceed the memory used by the model weights. For instance, caching keys/values for tens of thousands of tokens may consume more GPU memory than the model itself, severely limiting the throughput of LLM inference. Attention mechanisms tend to focus disproportionately on a few critical tokens, while many other tokens contribute very little to the output, suggesting not all past tokens are equally important for generating the next tokens. This insight has led to the development of several KV cache compression algorithms that evict less important tokens from the cache, thereby improving the throughput. LLMs can approximate full-context attention with minimal loss in accuracy by attending only to important tokens. \looseness=-1 

Several KV Cache compression methods have been proposed to manage KV cache growth without retraining the model. For example, StreamingLLM \cite{xiao2023efficient} focuses on retaining recent tokens or specific combinations of initial and recent tokens to sustain performance over longer contexts. Dynamic policies like H2O \cite{zhang2023h2o} utilize runtime information, such as attention weights, to estimate token importance. However, most of these eviction techniques focus on the trade-off between the compression achieved by eviction and the accuracy loss while ignoring the practical aspects of GPU memory management. Other approaches such as Scissorhands \cite{liu2023scissorhands} set a fixed budget of cached tokens and evict tokens whenever the budget is exceeded. These eviction methods free up memory to allow longer contexts and larger batches while preserving model quality as much as possible. While these methods can effectively minimize the GPU memory, they often require custom modifications to the LLM serving frameworks or storing attention weights. \looseness=-1

PagedAttention \cite{kwon2023efficient} is a memory management technique designed to address the inefficiencies in serving LLMs by optimizing how the KV cache is stored and accessed during inference. Traditional LLM serving frameworks allocate large contiguous memory spaces for the KV cache. This approach leads to substantial memory waste due to both internal and external fragmentation as memory chunks are over-reserved and scattered gaps prevent efficient reuse, especially when different requests generate variable sequence lengths. PagedAttention solves this issue by partitioning the KV cache into small, fixed-size blocks/pages rather than storing in a single contiguous chunk, each sequence’s KV cache is divided into many equally sized blocks, where the pages are only allocated on demand, reducing wasted memory. It is a widely used technique in production-ready runtimes such as vLLM to mitigate memory fragmentation.\looseness=-1

The current KV Cache eviction methods, which rely on attention scores (Q*K$^{T}$), cannot be integrated into PagedAttention, as FlashAttention never returns the attention score during the inference process. Also, the existing compression methods do not consider the paged structure in the vLLM framework while evicting tokens. They evict tokens across different pages, leading to a memory fragmentation issue for which PagedAttention was developed. Without synergy between token eviction and PagedAttention, the benefits of both approaches are limited. To address this challenge, we develop PagedEviction, a structured block-wise token eviction strategy optimized for a PagedAttention strategy. The high-level goal of PagedEviction is to limit KV cache by evicting entire blocks of tokens, so that memory usage (and computation per token) remains low, while the model’s accuracy on long-context tasks remains essentially unchanged from using a full cache. Unlike attention-score-based methods, PagedEviction does not require storing the attention weights during the forward pass. Our method relies on Key and Value tensors to reduce the KV Cache and requires no changes to the CUDA attention kernels. PagedEviction can free entire blocks of KV memory at once, avoiding the fragmentation or overhead that token-level evictions might introduce.\looseness=-1


    

\noindent\textbf{Contributions.} We propose \textbf{PagedEviction}, a novel and structured KV cache pruning algorithm designed for vLLM’s \textit{PagedAttention}. Our key contributions are as follows:

\begin{itemize}
    \item \textbf{Block-Aligned KV Cache Compression:} We introduce a structured block-wise KV cache eviction algorithm that aligns with vLLM's block-based memory. Our method computes token or block importance using a proxy for attention scores, avoiding the need to store attention, which is not possible with FlashAttention.

    \item \textbf{Eviction Strategy for Prefill and Decode:} PagedEviction operates in both prefill and decode phases. During prefill, we evict tokens based on per-token importance before block partitioning. During decode, we evict ab entire block only after the most recent block becomes full, reducing fragmentation and per-step eviction overhead.

    \item \textbf{High Accuracy Under Tight Budgets:} On LongBench datasets, PagedEviction consistently outperforms other attention-free baselines. For example, on GovReport with LLaMA-3.2-1B, it achieves a ROUGE score of 24.5 at a 1024-token budget, outperforming StreamingLLM (21.0) and KeyDiff (21.2) by 15–20\%. On MultiNews with LLaMA-3.2-3B, it achieves 23.6 ROUGE, outperforming Inverse Key L2-Norm by 1.1 points.

    \item \textbf{Hardware Benefits:} At a cache budget of 1024, PagedEviction achieves up to 3020 tokens/sec on LLaMA-1B, a \textbf{37\%} improvement over Full Cache (2200 tokens/sec), and a \textbf{39\%} improvement over Inverse Key L2-Norm (2170 tokens/sec). PagedEviction reduces latency by approximately \textbf{10--12\%} across 1B, 3B, and 8B models and scales robustly with larger model sizes. Its block-level evictions avoid frequent cache updates, making it more scalable under batch inference.
    
\end{itemize}


\begin{figure*}
    \centering
     \includegraphics[width=\linewidth]{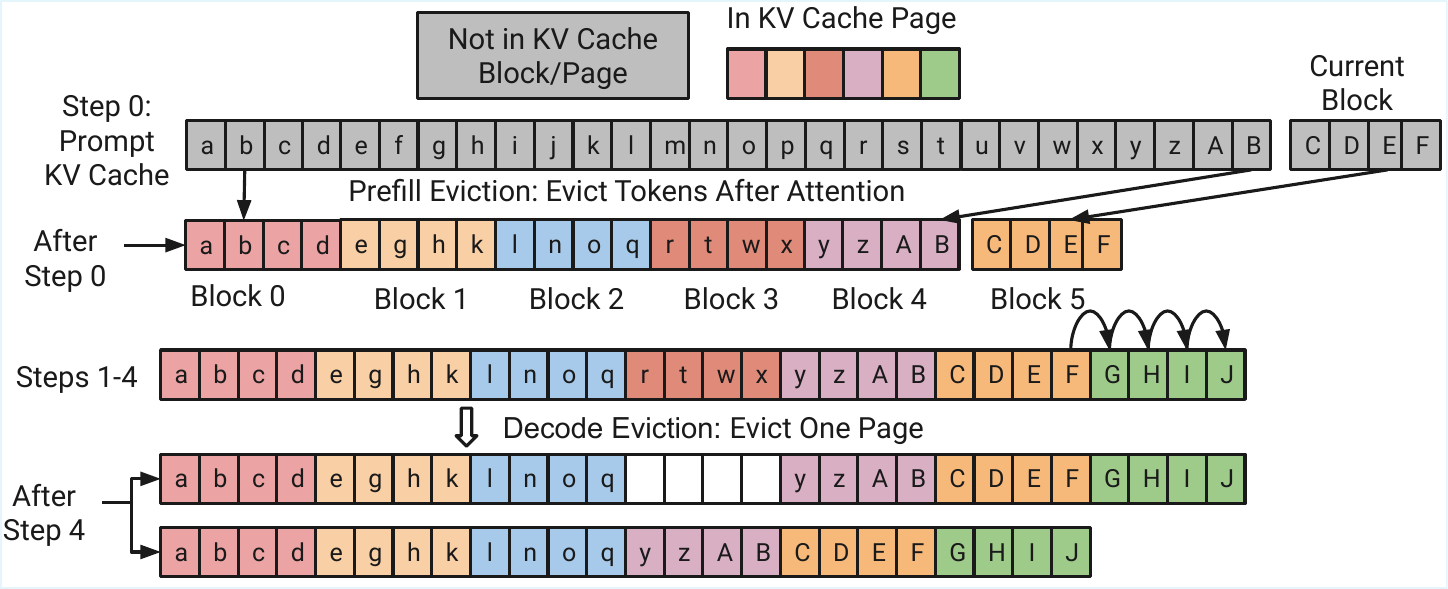}
        \caption{Illustration of the PagedEviction technique integrated into vLLM with a block size ($B$) and a cache budget ($C$). During the \textit{prefill} stage, after the initial set of Key and Value tokens is generated, a subset of tokens is evicted until the cache budget is reached. In the \textit{decode} phase, one block of tokens is evicted once the recent block becomes full.}    
        \vspace{-4mm}
        \label{fig:paged_compression}
    \captionsetup{justification=centering}
\end{figure*}

\section{Background and Related Work}

\subsection{Self-Attention}

Given the input tensor $\mathbf{X}$, the multi-head attention mechanism performs multiple attention operations in parallel. It computes three projections: Query ($\mathbf{Q}$), Key ($\mathbf{K}$), and Value ($\mathbf{V}$) matrices through three linear layers by multiplying $\mathbf{X}$ with $\mathbf{W}_Q, \mathbf{W}_K, \mathbf{W}_V \in \mathbb{R}^{d \times d_k}$. The attention activation is calculated using the scaled dot-product attention. This process is repeated $H$ times, each with different learned projections $\mathbf{W}_Q^{(i)}, \mathbf{W}_K^{(i)}, \mathbf{W}_V^{(i)}$ for each head $h$. The outputs from different heads are concatenated and projected back to the original dimension $d$ using a final learned matrix $\mathbf{W}_O \in \mathbb{R}^{hd_k \times d}$.\looseness=-1



\subsection{KV Cache}
During the autoregressive LLM inference, the tokens are generated sequentially. Without a caching mechanism, the Key ($\mathbf{K}$) and Value ($\mathbf{V}$) states are computed in every step for all the previously predicted tokens. The KV Cache addresses this inefficiency by saving Key and Value activations for each token. Instead of recalculating these activations, the model retrieves the cached $\mathbf{K}$ and $\mathbf{V}$ activations and concatenates with the current token. During the current step, the attention computation is performed as per Eq \ref{eq:attn_kv_cache}. 
\begin{equation}\label{eq:attn_kv_cache}
\text{Attention}(\mathbf{Q}_t, [\mathbf{K}_{1:t-1}; \mathbf{K}_t], [\mathbf{V}_{1:t-1}; \mathbf{V}_t])
\end{equation}
where $[;]$ denotes concatenation along the sequence dimension, and $\{\mathbf{K}_{1:t-1}, \mathbf{V}_{1:t-1}\}$ are retrieved from memory. 
Although KV caching can significantly reduce computational costs by avoiding redundant calculations, storing the cached values for every token in the sequence incurs substantial memory, which grows linearly with the sequence length and batch sizes. For a model with $L$ layers, $H$ attention heads, $d$ hidden size per head, and a sequence length of $S$, the memory required is 2 $\times$ S $\times$ $L$ $\times$ $H$ $\times$ $d$ $\times$ 2 bytes.\looseness=-1 

\subsection{KV Cache Eviction}
The KV Cache Eviction methods identify less important tokens in $K[1 : t-1]$ and $V[1 : t-1]$ by developing a function $fun_{kv}$ that identifies redundant tokens. The attention operation ($\text{Att}(\mathbf{Q}, \mathbf{K'}, \mathbf{V'})$) is performed on non-evicted Key and Value tokens. Several strategies have been developed to evict less important tokens based on their importance and relevance. H2O \cite{zhang2024h2o} uses cumulative normalized attention scores to retain high-impact tokens while preserving recent tokens due to their strong correlations with current tokens. StreamingLLM \cite{xiao2023efficient} maintains a limited number of initial KV pairs to ensure model performance remains stable. Keyformer \cite{adnan2024keyformer} addresses the issue of token removal by introducing regularization techniques to smooth and approximate the original softmax probability distributions, mitigating distortions caused by token eviction. PyramidInfer \cite{yang2024pyramidinfer} dynamically adjusts KV cache size layer-wise based on redundancy. However, these methods depend on attention scores for eviction, requiring CUDA kernel modifications to track them. 
\looseness=-1



\subsection{PagedAttention: Blocked KV Caching}

PagedAttention leverages a block-based structure for KV cache management inspired by OS virtual memory management. It dynamically allocates memory for KV tokens per attention head. Logical KV blocks map to physical blocks via a block table to minimize memory fragmentation. Dynamic allocation ensures efficient handling of variable sequence lengths,  eliminating the need for pre-allocating large memory spaces in advance. \looseness=-1



\section{Challenges and Limitations of SOTA}

In this section, we outline the key limitations of existing KV cache eviction methods and present our proposed solutions to address them.

\textbf{Limitation 1: KV Cache Memory Organization.} 
vLLM structures the KV cache into blocks, each storing a fixed number of key-value tokens per attention head. However, existing eviction methods do not consider this block structure and often evict different number of tokens across different blocks. This leads to block fragmentation, disrupting the uniformity of the memory layout. Such fragmentation reduces memory allocation efficiency for future requests, as blocks no longer maintain consistent occupancy.\looseness=-1 

\textbf{Our Solution.} 
We introduce a block-aware structured eviction mechanism (Figure~\ref{fig:paged_compression}) that preserves the native block alignment in vLLM after eviction. This approach maintains structural consistency, reduces fragmentation, and improves memory reuse efficiency.

\textbf{Limitation 2: Block-Aware Eviction.} 
Most existing methods have been designed for HuggingFace-based implementations, where KV cache tokens are stored contiguously in memory. This simplifies global token importance comparisons and merging of retained tokens. However, in vLLM, global comparisons are computationally prohibitive due to its non-contiguous block-based storage. Evaluating importance across all tokens during decoding introduces significant overhead, especially for long sequences.\looseness=-1 

\textbf{Our Solution.} 
We restrict eviction to block-level granularity during decoding by computing a single score per block rather than per token. This eliminates the need to move tokens across blocks and allows for efficient full-block eviction without modifying the underlying attention kernel. Our method retains consistent block sizes throughout inference and enhances memory reuse by preserving the structural layout of evicted blocks.

\textbf{Limitation 3: Dependence on Attention Scores.} 
Methods like H2O~\cite{zhang2023h2o} rely on cumulative attention scores to identify unimportant tokens. However, optimized attention kernels such as FlashAttention do not return attention scores. Accessing or maintaining attention scores during inference requires integration with schedulers or host memory, which is decoupled from the CUDA kernel in vLLM. This dependency introduces substantial runtime and memory overhead, undermining inference efficiency.\looseness=-1 

\textbf{Our Solution.} 
We avoid reliance on attention scores by using a proxy importance metric derived solely from the static key and value states already stored in the KV cache. This proxy is computed on-the-fly without modifying the attention kernel or maintaining additional memory, ensuring compatibility with optimized inference paths.\looseness=-1 

\textbf{Limitation 4: Per-Step Token Eviction.} 
Eviction strategies such as StreamingLLM~\cite{xiao2023efficient} and H2O~\cite{zhang2024h2o} perform eviction at every decoding step, requiring constant updates to the KV cache table across all layers. This results in significant per-step latency and reduces throughput, especially under large batch inference.\looseness=-1 

\textbf{Our Solution.} 
We adopt a coarse-grained eviction strategy wherein entire blocks (pages) are evicted only when the current block reaches capacity. This reduces the frequency of eviction operations and avoids the creation of partially filled blocks, which are incompatible with vLLM. Our approach simplifies eviction logic and reduces runtime overhead while maintaining efficient cache utilization.\looseness=-1

\section{PagedEviction}\label{PagedEviction}

Our PagedEviction method is divided into two different components: Prefill KV cache eviction and Decode KV Cache eviction. This design choice is to align with the implementation in the vLLM framework. Figure \ref{fig:paged_compression} summarizes our PagedEviction algorithm in both prefill and decode phases. 

\subsection{Token Importance} \label{token_importance}
vLLM's attention kernel, implemented in CUDA, is separate from its KV Cache memory management. Methods like H2O \cite{zhang2024h2o} require significant changes to the attention kernel to maintain a running sum of attention scores per head and per layer across different blocks. Therefore, we need a cost-effective method to evaluate the token importance without adding complexity and additional memory. The value activation matrix ($V$) represents the feature embeddings of each token that will be selectively combined to create the contextualized output. The Value tensor ($V$) holds the actual information to extract relevant pieces of information from the hidden states. This allows LLM to focus on specific aspects of the input that are most important in the current context. Also, previous work by Devoto et al. \cite{devoto2024simple} observed a unique correlation between the Key tensor and attention weights. The L2-norm of each Key token is inversely proportional to the cumulative attention score of each token. Therefore, we compute the score/importance ($S$) of each token or block using the ratio of the L2 norm of the Value token to that of the Key token ($\|V_i\|_2 \div \|K_i\|_2$). This requires us to fetch token importance directly from the static KV cache, eliminating the need to rely on the cumulative attention score.\looseness=-1  

\begin{algorithm}[h]
\caption{Token/Block Importance} 
\label{alg:token_block_Importance}
\begin{algorithmic}[1]
\Require Key $K$, Value $V$, Page Size $B$, Eviction $\mathsf{M}$ 
\If{$\mathsf{M} = \texttt{token}$} 
    \For{each token $i$}
        \State $S_i \gets \|V_i\|_2 \,/\, \|K_i\|_2$
    \EndFor
\ElsIf{$\mathsf{M} = \texttt{block}$}
    \State Divide $K$ \& $V$ into blocks of size $B$
    \For{each block $j$}
        \State Compute block score $S_j$ as the average:
        \State \quad $S_j \gets \frac{1}{B} \sum\limits_{i \in \text{block } j} \left( \|V_i\|_2 \,/\, \|K_i\|_2 \right)$
    \EndFor
\EndIf

\end{algorithmic}
\end{algorithm}


\subsection{PagedEviction: Prefill Phase}

During the prefill phase, the prompt tokens are processed in a single forward pass through multiple layers, where Q, K, and V matrices are generated in each layer. In each layer, the contiguous Key and Value tensors are written to the corresponding KV cache blocks. We implement token eviction before the Key and Value states are divided into different pages. This approach is important because performing the eviction after storing tokens across different blocks would require significant memory reordering and movement between blocks. Therefore, during the prefill phase, PagedEviction performs token-level KV cache compression by computing the importance of each token and evicting the least important ones to reach a fixed cache budget ($B$). As shown in Algorithm~\ref{alg:paged_compression_prefill}, the process begins with a forward pass over the prompt to generate the initial KV Cache. Following the self-attention operation, PagedEviction computes a per-token importance score $S_i = \|V_i\|_2 / \|K_i\|_2$, where tokens with lower scores are considered less critical to future predictions. We evict $E$ tokens ($E$ = $L$ - $C$) from the input length ($L$). This simple yet effective scoring mechanism enables PagedEviction to compress the KV cache before decoding begins, reducing memory usage while preserving important KV states.




\begin{algorithm}[h]
\caption{PagedEviction: Prefill Token Compression} 
\label{alg:paged_compression_prefill}
\begin{algorithmic}[1]
\Require Input Hidden State $I$, Cache Budget $C$ and Block/Page Size $B$ 
\State \textbf{Initialization:} No KV cache exists initially
\State Process the prompt to generate initial KV cache 
\For{One Forward Pass}
    \State $Q = Q_{w}(I), K = K_{w}(I), V = V_{w}(I)$ 
    \State Attn = Self-Attention($Q, K, V$)
    \State $O$ = Out$_{proj}$(Attn)
    \State Compute Importance of token $i$ ($S_i$) 
    \State $S_i$ = $||V_i||_2$ $\div$ $||K_i||_2$
    \State $L$ = Sequence Length of Input $I$
    \State Evict Tokens ($E$) = $L$ - $C$
    \State Evict $E$ Tokens in Key and Value Cache
    \State $K, V$ = Evict($K$, $E$), Evict($V$, $E$)
    \State Divide K and V into different pages based on $B$  
\EndFor
\end{algorithmic}
\end{algorithm}


\subsection{PagedEviction: Decode Phase}

During the decode phase, PagedEviction evicts one KV cache page/block after every recent block is completely full. As shown in Algorithm~\ref{alg:paged_compression_decode_percentage_eviction} and illustrated in Figure~\ref{fig:paged_compression}, after each newly generated block of tokens (i.e., when the current sequence length $L$ is a multiple of the block size $B$), the algorithm evaluates all existing pages in the cache. Each page is assigned an importance score based on the aggregated token-level scores within that page, which is the mean of the $\|V_i\|_2 / \|K_i\|_2$ ratio across all tokens $i$ in the block. The page with the lowest importance score is evicted from the cache, and the internal KV cache block table is updated accordingly. This strategy ensures that PagedEviction continuously frees memory in a structured, block-wise fashion, enabling long-context inference with minimal disruption to the model's performance.  We evict a block only when the last block becomes full. This strategy ensures that we maximize space utilization, as evicting a single token does not free the entire block, making partial eviction inefficient. Additionally, triggering eviction only when a new block is needed reduces the overall frequency of eviction operations. This process of generating new tokens and periodically evicting less important ones continues until either the end-of-sequence (EOS) token is reached or the maximum number of new tokens is generated. In this way, we preserve the most significant information for ongoing token generation while maintaining the block structure in vLLM's PagedAttention. Our proposed eviction policy offers several advantages over random token eviction strategies. We eliminate the need for extensive token rearrangement across all blocks during each eviction step as we evict a fixed number of tokens equal to the original block size. Also, the attention kernel can operate without modifications, as all blocks maintain a uniform size throughout the process, and our method is compatible with Flash Attention.

\begin{algorithm}[h]
\caption{PagedEviction: Decode Phase Compression}
\label{alg:paged_compression_decode_percentage_eviction}
\begin{algorithmic}[1]
\Require Input Hidden State $I$, Block/Page Size $B$, Cache Budget $C$ and KV Cache KV$_{Cache}$

\While{EOS or max new tokens}
    \State $L$ = Current Sequence Length 
    \If{$L$ \% $B$ == 0}
        \State $N$ = Number of KV Cache Pages/Blocks  
        \State Compute Score of Page $i$ ($S_i$)  
        \State $S_i$ = $||V_i||_2$ $\div$ $||K_i||_2$   
        \State Evict One page with the lowest score $S_i$
        \State $K, V$ = Evict\_page($K$), Evict\_page($V$)
        \State Update the KV Cache Block Table 
    \EndIf
    \State Attn = PagedAttention($Q, K, V$)
    \State $O$ = Out$_{proj}$(Attn)
    \State Continue generating new tokens
\EndWhile
\end{algorithmic}
\end{algorithm}

\begin{figure*}[t]
    \centering
    \includegraphics[width=0.8\textwidth]{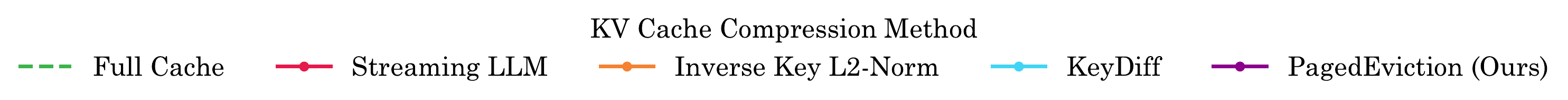}\\[-0.3ex]
    \begin{subfigure}[b]{0.195\textwidth}
        \centering
        \includegraphics[width=\linewidth]{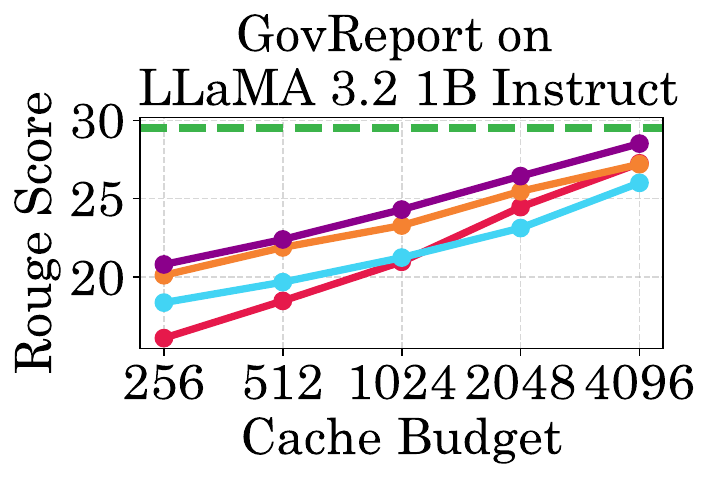}
    \end{subfigure}
    \begin{subfigure}[b]{0.195\textwidth}
        \centering
        \includegraphics[width=\linewidth]{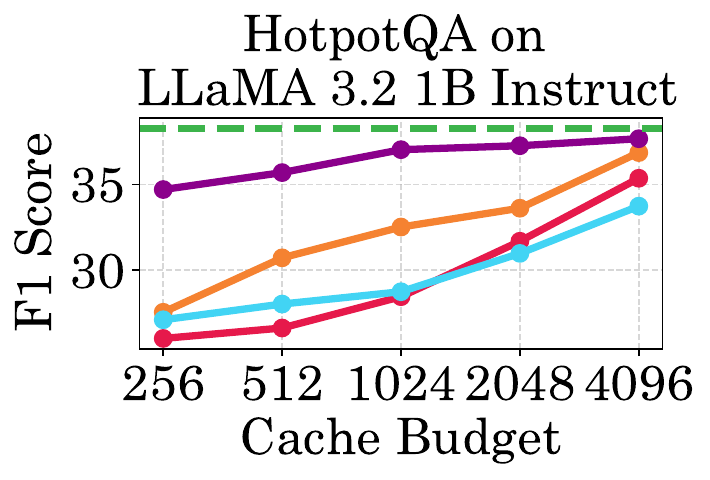}
    \end{subfigure}
    \begin{subfigure}[b]{0.195\textwidth}
        \centering
        \includegraphics[width=\linewidth]{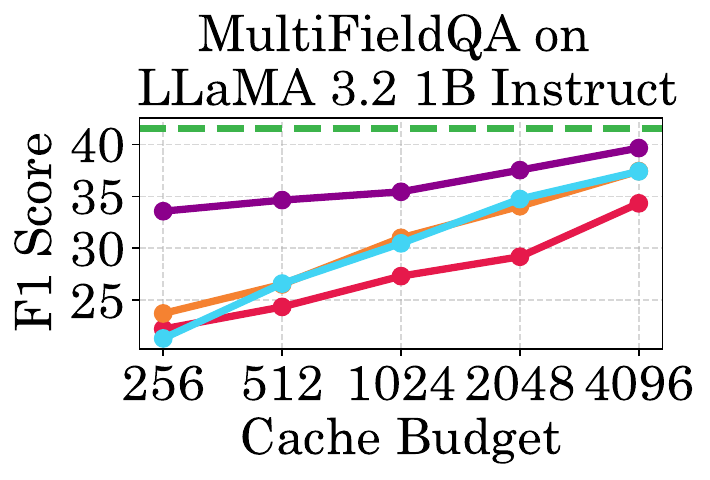}
    \end{subfigure}
    \begin{subfigure}[b]{0.195\textwidth}
        \centering
        \includegraphics[width=\linewidth]{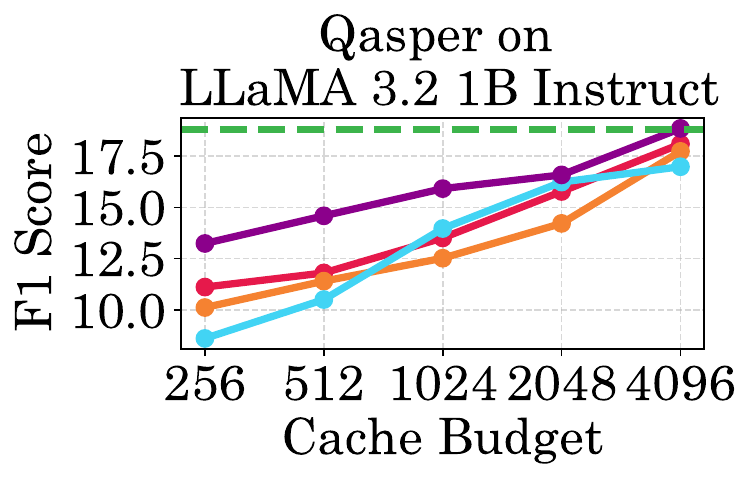}
    \end{subfigure}
    \begin{subfigure}[b]{0.195\textwidth}
        \centering
        \includegraphics[width=\linewidth]{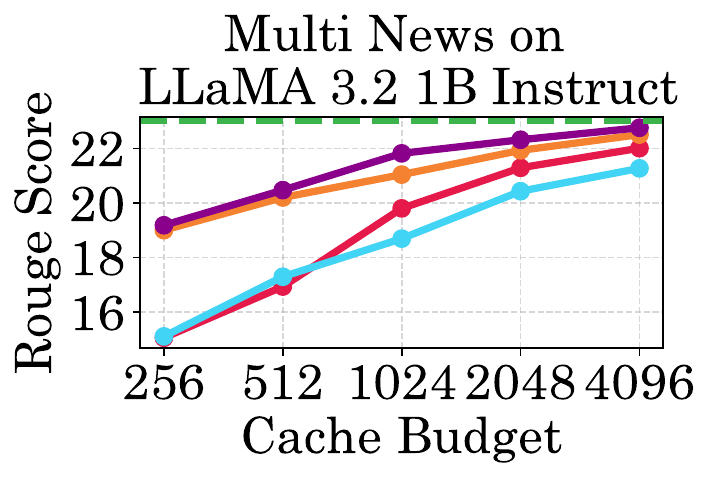}
    \end{subfigure}
    \begin{subfigure}[b]{0.195\textwidth}
        \centering
        \includegraphics[width=\linewidth]{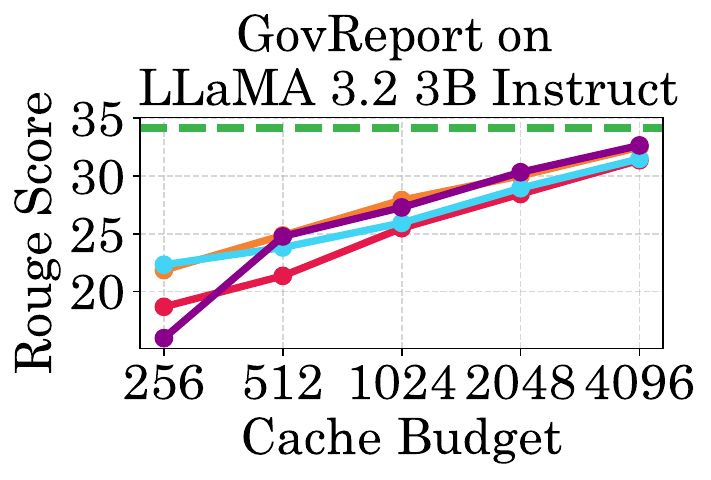}
    \end{subfigure}
    \begin{subfigure}[b]{0.195\textwidth}
        \centering
        \includegraphics[width=\linewidth]{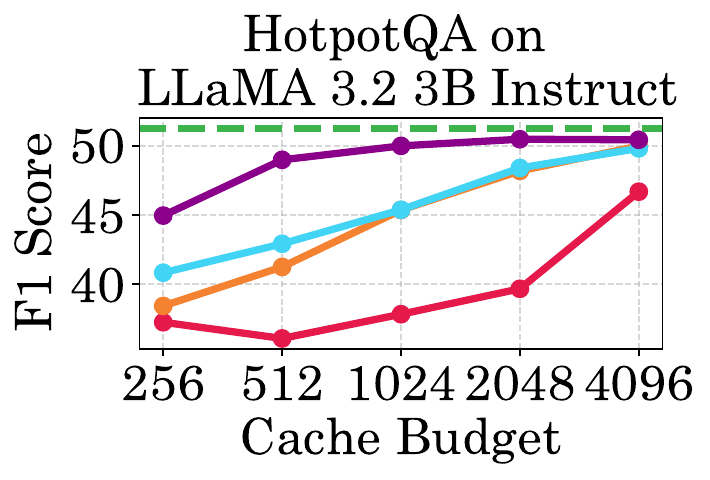}
    \end{subfigure}
    \begin{subfigure}[b]{0.195\textwidth}
        \centering
        \includegraphics[width=\linewidth]{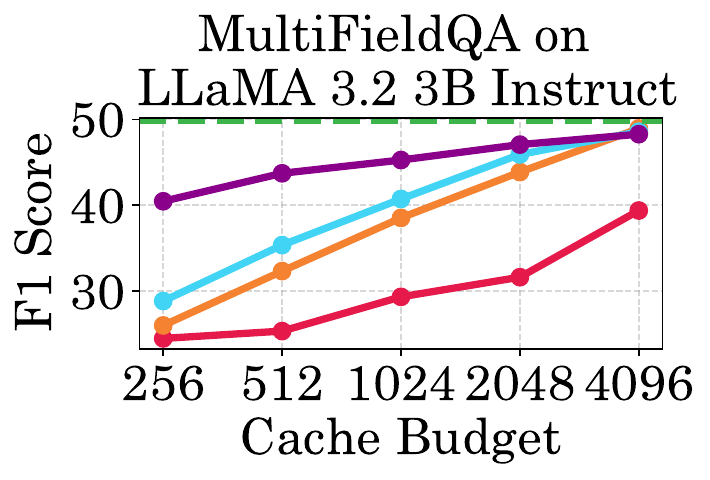}
    \end{subfigure}
    \begin{subfigure}[b]{0.195\textwidth}
        \centering
        \includegraphics[width=\linewidth]{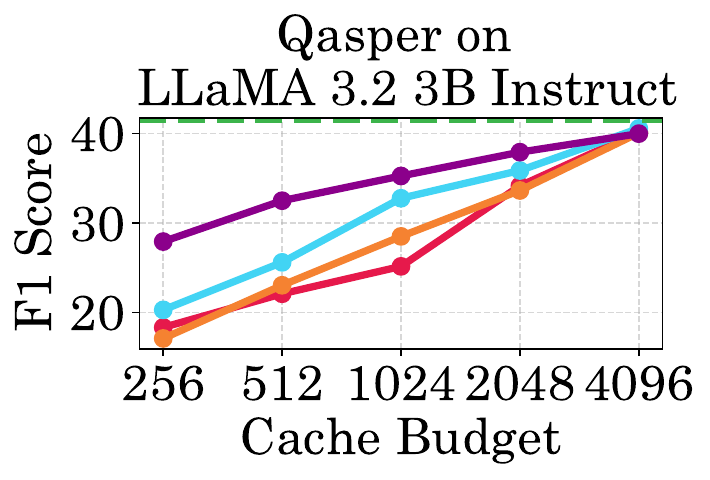}
    \end{subfigure}
    \begin{subfigure}[b]{0.195\textwidth}
        \centering
        \includegraphics[width=\linewidth]{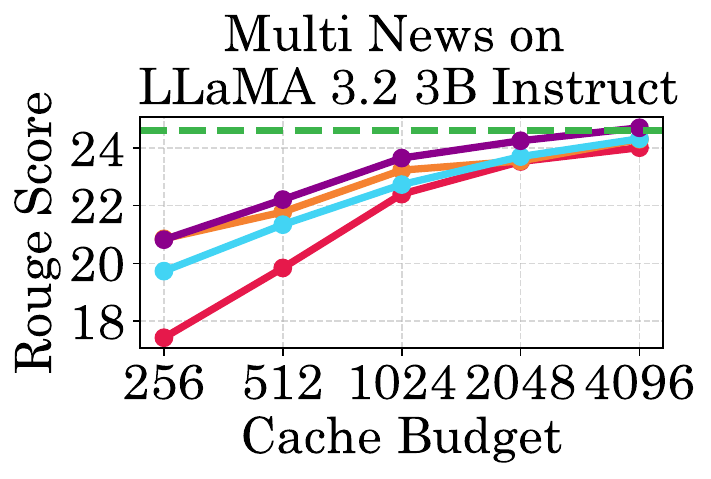}
    \end{subfigure}
    \begin{subfigure}[b]{0.195\textwidth}
        \centering
        \includegraphics[width=\linewidth]{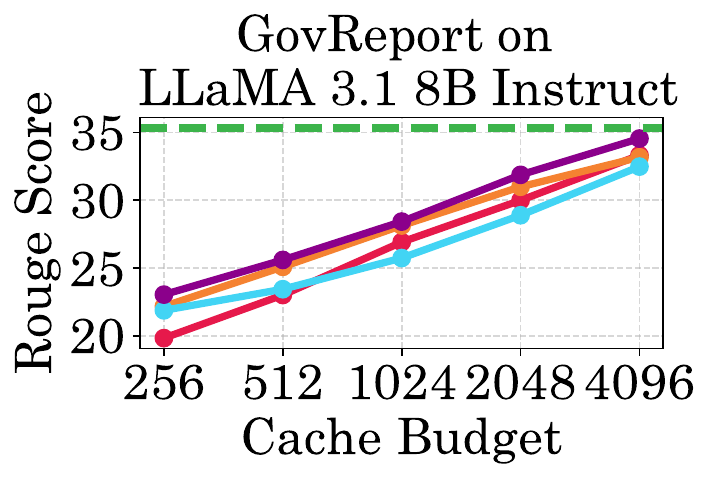}
    \end{subfigure}
    \begin{subfigure}[b]{0.195\textwidth}
        \centering
        \includegraphics[width=\linewidth]{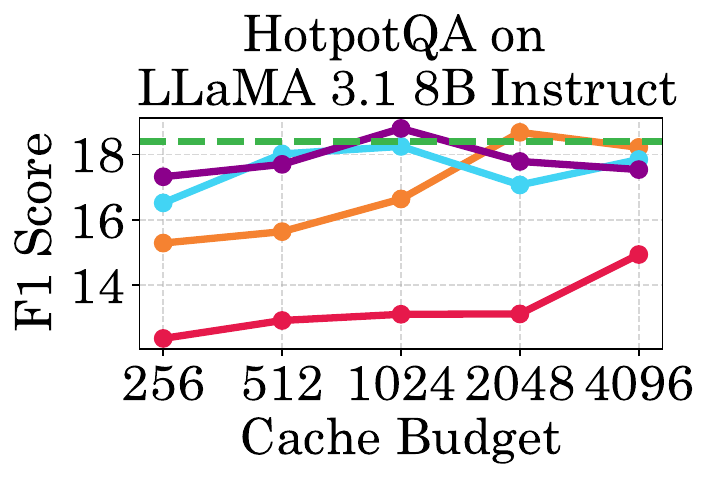}
    \end{subfigure}
    \begin{subfigure}[b]{0.195\textwidth}
        \centering
        \includegraphics[width=\linewidth]{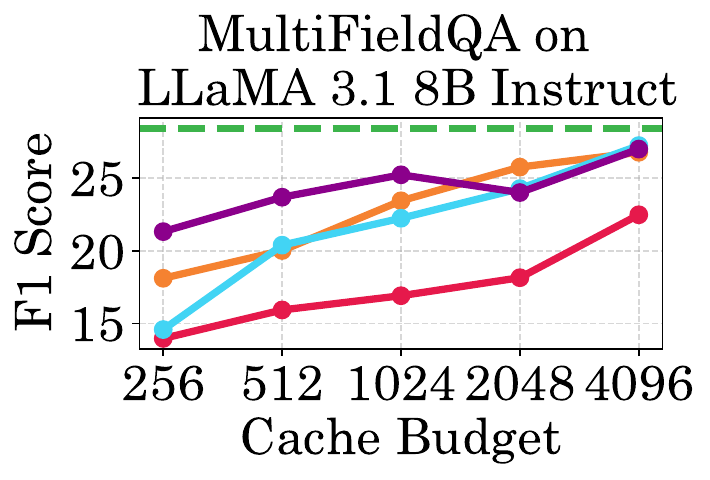}
    \end{subfigure}
    \begin{subfigure}[b]{0.195\textwidth}
        \centering
        \includegraphics[width=\linewidth]{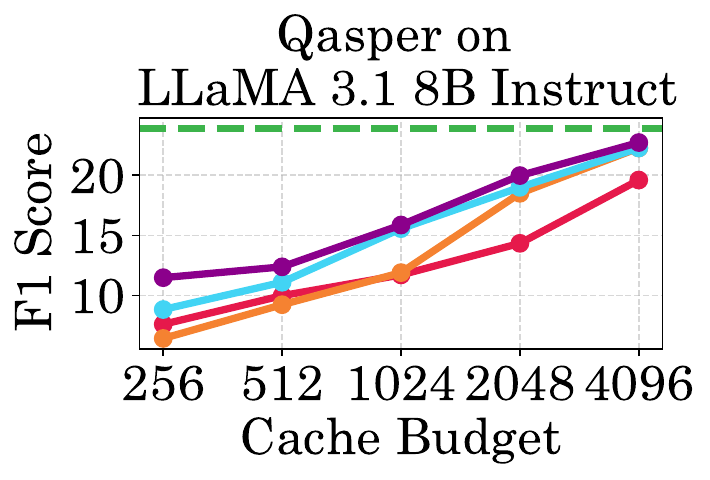}
    \end{subfigure}
    \begin{subfigure}[b]{0.195\textwidth}
        \centering
        \includegraphics[width=\linewidth]{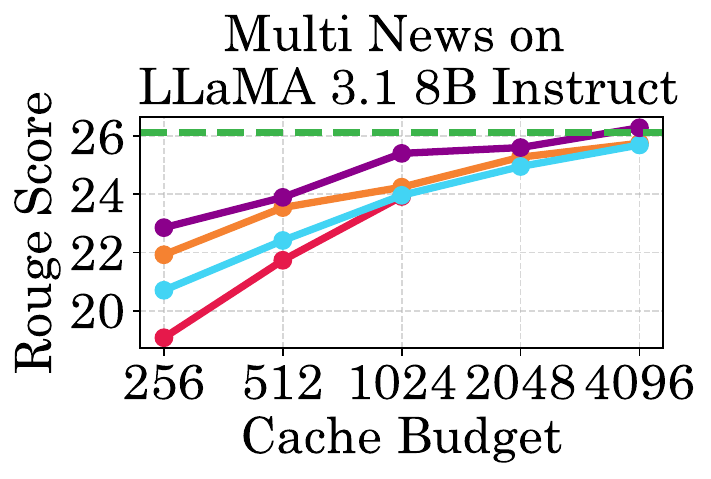}
    \end{subfigure}
    \caption{Accuracy vs Cache Budget of 1B, 3B, and 8B models on GovReport, HotpotQA, MultiFieldQA, Qasper and Multi News Datasets. PagedEviction consistently achieves higher or comparable accuracy relative to other attention-free baselines, especially under low cache budgets (e.g., 256–1024), and closely approaches Full Cache performance at high budgets (2048–4096)}
    \label{fig:longbench_results}
\end{figure*}

\section{Evaluation, Results and Discussion}

\subsection{Experimental Setup}
We evaluate our method under varying memory constraints using cache budgets of {256, 512, 1024, 2048, 4096}. For all experiments, we use a page size of 16, which has been shown to be optimal for vLLM \cite{kwon2023efficient}. Nonetheless, our proposed PagedEviction algorithm is compatible with any page size and cache budget configuration. We implement our method on top of vLLM version 0.9.0. To assess the effectiveness of our approach, we adopt several datasets from the LongBench benchmark \cite{bai2023longbench}, including HotpotQA \cite{yang2018hotpotqa}, Qasper \cite{dasigi2021dataset}, GovReport \cite{huang2021efficient}, MultiNews \cite{fabbri2019multi}, and MultiFieldQA. These datasets have long input context and long outputs. We exclude simple QA tasks like 2WikiMultihopQA \cite{ho2020constructing}, whose output is one or two tokens. All experiments are conducted on NVIDIA A100 GPUs with 40GB memory. For throughput evaluation, we use synthetic inputs with the following setup: input sequence length of 1024 tokens, output sequence length of 8192 tokens, and 64 concurrent requests per batch. We report throughput (total number of input + output tokens processed per second) and Time per Output Token (TPOT). We evaluate our method on three LLMs of different scales, which are Meta-Llama-3.2-1B-Instruct, Meta-Llama-3.2-3B-Instruct, and Meta-Llama-3.1-8B-Instruct \cite{meta2024llama3} to demonstrate generality across model sizes.\looseness=-1

\subsection{Baseline Methods}
We evaluate our proposed PagedEviction algorithm against strong baselines that do not rely on attention scores to estimate token importance. To the best of our knowledge, the most relevant such baselines include: Full Cache (no eviction), StreamingLLM \cite{xiao2023efficient}, Inverse Key L2-Norm \cite{devoto2024simple}, and KeyDiff \cite{park2025keydiff}. These methods estimate token importance using static representations, such as key or value states, without modifying the core attention mechanism. StreamingLLM maintains a fixed-size KV cache consisting of a small prefix of initial tokens (e.g., the first 4 tokens) that are retained as "attention sinks", combined with a sliding window of the most recent tokens (e.g., the last 1024 tokens).
Inverse Key L2-Norm prunes the cache by evicting tokens with high L2 norms of their key vectors, under the assumption that lower-norm keys are less influential. KeyDiff emphasizes key vector diversity by evicting tokens whose keys are redundant, i.e., highly similar to others, thereby preserving a more diverse and informative KV cache. We categorize attention-free KV cache eviction methods into structured and unstructured approaches. Structured methods, such as our PagedEviction and StreamingLLM, evict tokens within a single block or remove entire blocks together. In contrast, unstructured methods like Inverse Key L2-Norm and KeyDiff operate at the token level, evicting tokens across the entire sequence based on individual importance scores. All of these baselines evict tokens at a fine-grained token level across different pages during each decoding step. Importantly, none of them require changes to the underlying CUDA attention kernel. Since our primary objective is to retain compatibility with vLLM and avoid modifying the core attention implementation, we restrict our evaluation to baselines that can be integrated within the vLLM runtime. Other methods, such as H2O \cite{zhang2024h2o}, require kernel modifications that hinder deployment efficiency and fall outside the scope of our work. Thus, for a fair and practical comparison, we compare PagedEviction against the three non-trivial but vLLM-friendly methods.\looseness=-1

\subsection{LongBench Results}

Figure \ref{fig:longbench_results} compares our PagedEviction with several attention-free KV cache compression methods across multiple datasets and models. We evaluate across three LLaMA variants (3.2 1B Instruct, 3.2 3B Instruct, and 3.1 8B Instruct) across varying cache budgets from 256 to 4096 tokens. Our PagedEviction method (highlighted in purple) demonstrates remarkably consistent performance across several settings, maintaining scores competitive with the full cache baseline while requiring substantially less memory footprint. The StreamingLLM approach shows moderate compression effectiveness, particularly excelling in scenarios with larger cache budgets, which aligns with its design principle of maintaining attention sinks for stable long-context processing. The Inverse Key L2-Norm method exhibits variable performance, with notable degradation on certain tasks like Qasper, suggesting that low-norm key embeddings may not universally correlate with attention importance across all model architectures and tasks. The KeyDiff method demonstrates steady but modest compression gains, reflecting its similarity-based eviction strategy that preserves geometrically distinctive keys. These results underscore the critical importance of task-specific and model-aware selection of KV cache compression techniques, as no single method universally dominates across all evaluation scenarios.\looseness=-1 

Across all models and datasets, PagedEviction consistently achieves superior performance under tight cache budgets, demonstrating its ability to retain high accuracy while significantly reducing memory usage. For instance, our PagedEviction on LLaMA-3.2-1B-Instruct on the GovReport dataset (long context summarization) consistently outperforms all baselines. At a cache budget of 1024, PagedEviction achieves a score of $\sim$24.5, which is $\sim$15–20\% higher than StreamingLLM ($\sim$21) and KeyDiff ($\sim$21.2). At the cache budget of 4096, it reaches $\sim$29.5, closely matching the full-cache score 30, demonstrating its effectiveness across all memory constraints. On the MultiNews dataset using the LLaMA-3.2-3B-Instruct model, PagedEviction achieves a ROUGE score of approximately $\sim$23.6 at a cache budget of 1024, outperforming Inverse Key L2-Norm ($\sim$22.5) and StreamingLLM ($\sim$22.0) by roughly 5–9\%. At the budget of 4096, our method nearly matches the full-cache performance ($\sim$24.5), demonstrating both high compression efficiency and minimal accuracy degradation. This highlights the effectiveness and generalizability of our block-wise eviction policy across tasks and models. Our main contribution is the block-wise KV cache eviction mechanism, which can potentially serve as a basis for further optimizations. Techniques such as layer-wise budget allocation \cite{yang2024pyramidinfer} and quantized KV caching \cite{dong2024qaq} can be built on top of our approach to further enhance performance.\looseness=-1


\begin{figure*}[t]
    \centering
    \includegraphics[width=0.9\textwidth]{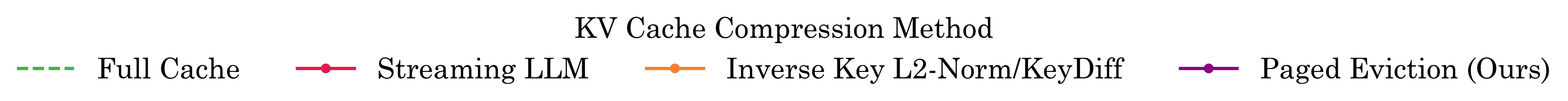}\\[-0.3ex]
    \begin{subfigure}[b]{0.22\textwidth}
        \centering
        \includegraphics[width=\linewidth]{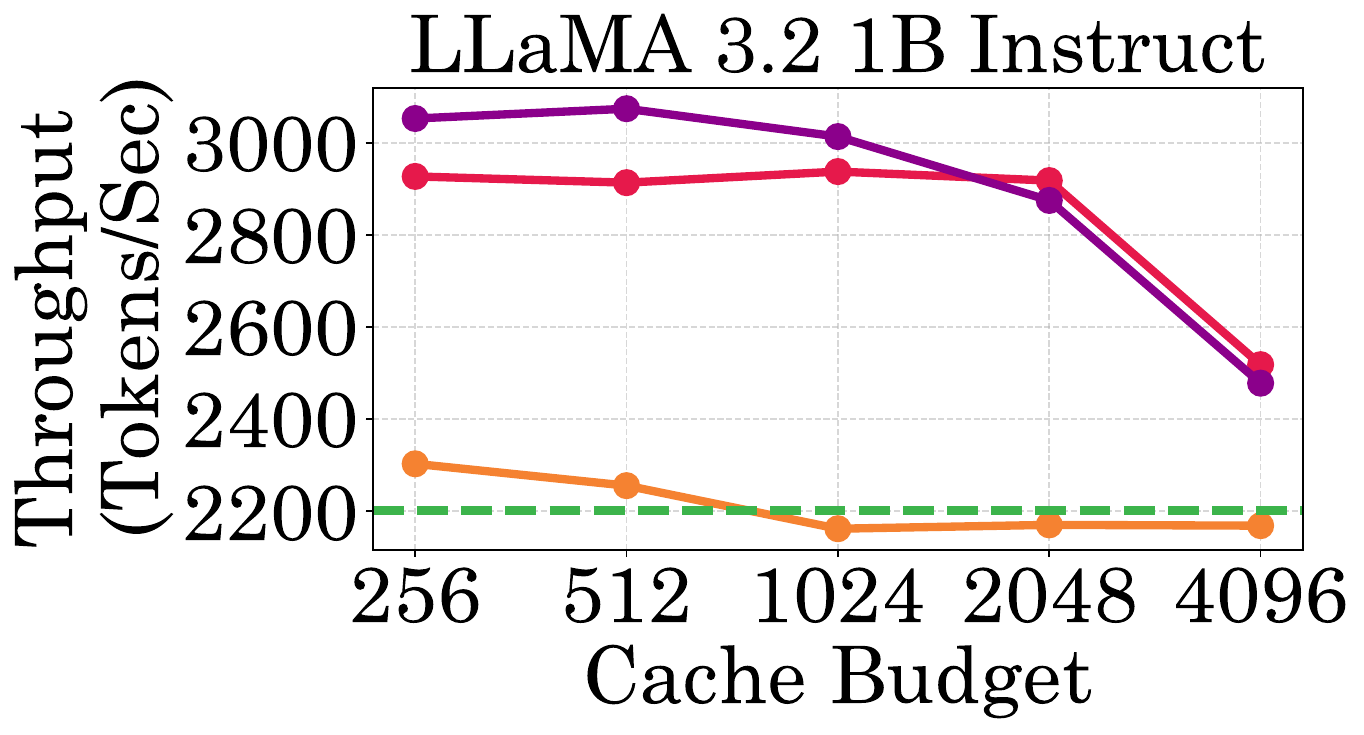}
        \caption{LLaMA-1B}
    \end{subfigure}
    \begin{subfigure}[b]{0.22\textwidth}
        \centering
        \includegraphics[width=\linewidth]{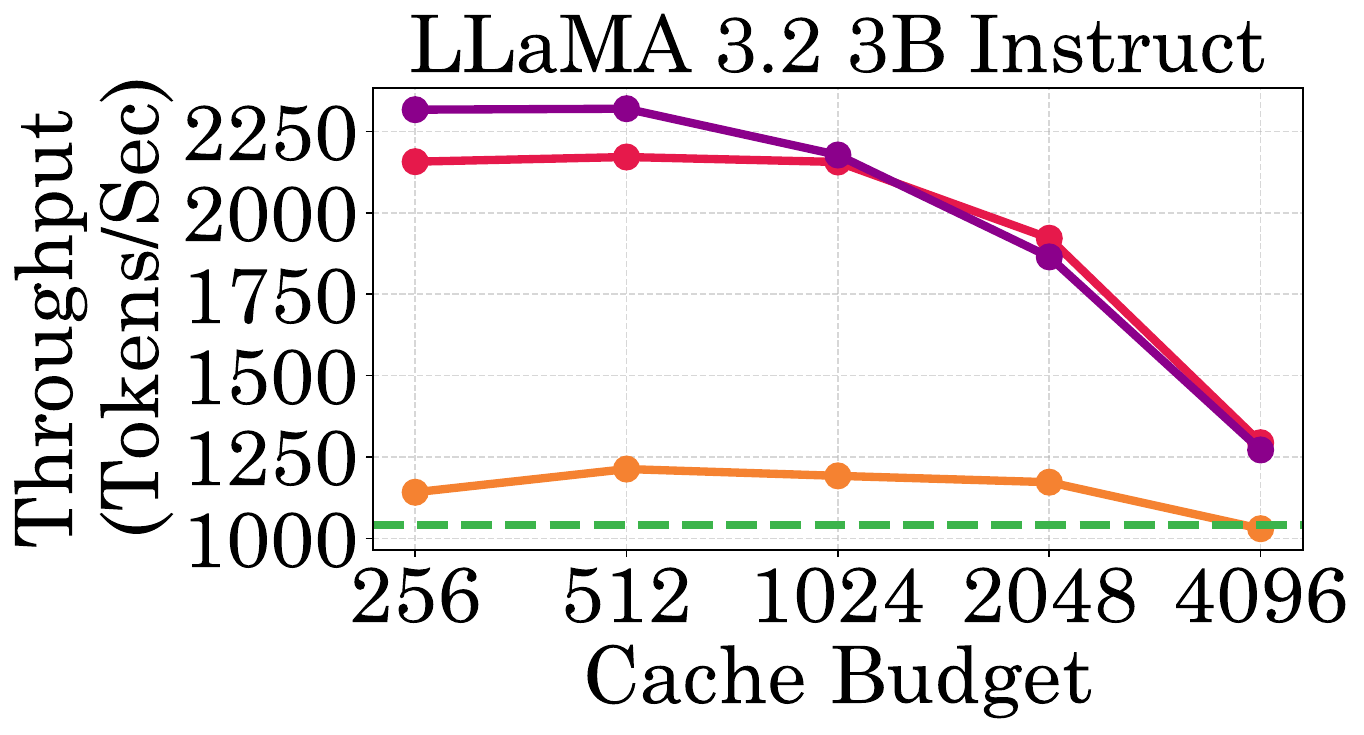}
        \caption{LLaMA-3B}
    \end{subfigure}
    \begin{subfigure}[b]{0.22\textwidth}
        \centering
        \includegraphics[width=\linewidth]{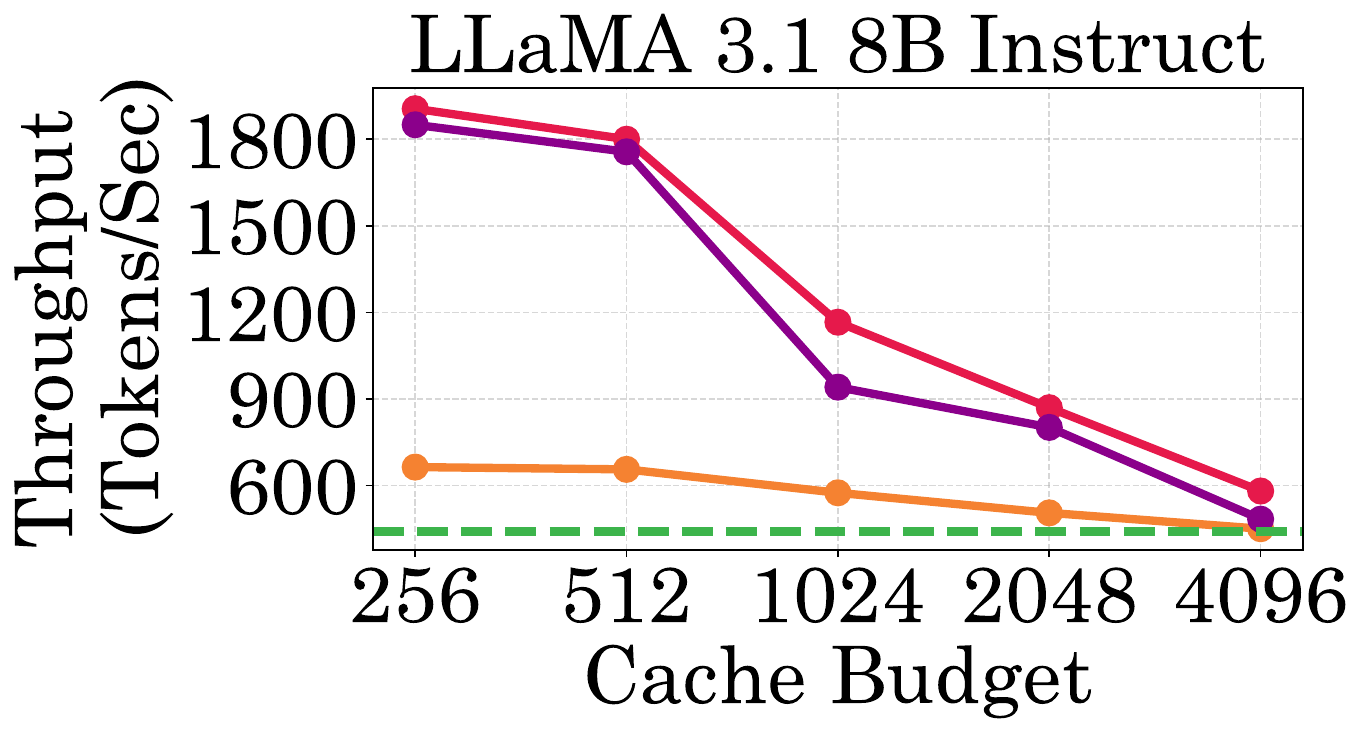}
        \caption{LLaMA-8B}
    \end{subfigure}
    \begin{subfigure}[b]{0.22\textwidth}
        \centering
        \includegraphics[width=\linewidth]{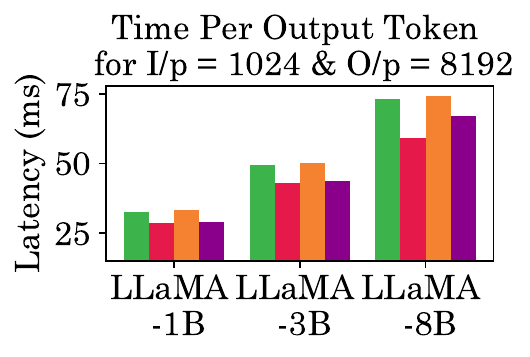}
        \caption{Time Per Output Token}
    \end{subfigure}
    \caption{(a,b,c) Cache Budget vs Throughput of Llama-3.2-1B-Instruct, Llama-3.2-3B-Instruct and Llama-3.1-8B-Instruct models and (d) Time Per Output Token of 1B, 3B and 8B models}
    \label{fig:throughput_latency}
\end{figure*}

\begin{figure*}[t]
    \centering
    \includegraphics[width=0.9\textwidth]{Figs/graphs/legend.pdf}\\[-0.3ex]
    \begin{subfigure}[b]{0.195\textwidth}
        \centering
        \includegraphics[width=\linewidth]{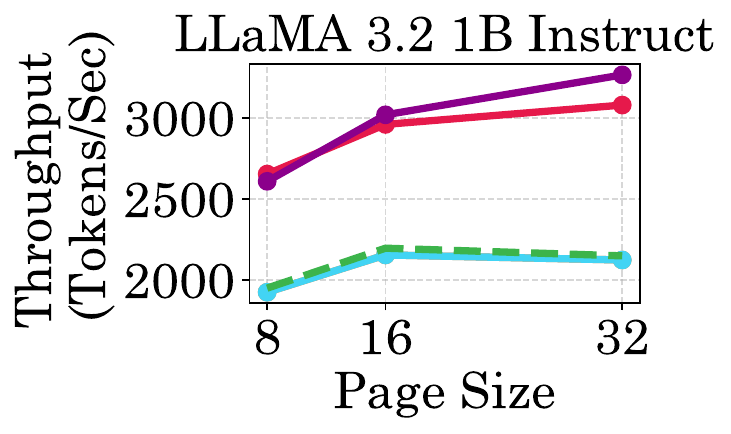}
        \caption{Throughput}
    \end{subfigure}
    \begin{subfigure}[b]{0.195\textwidth}
        \centering
        \includegraphics[width=\linewidth]{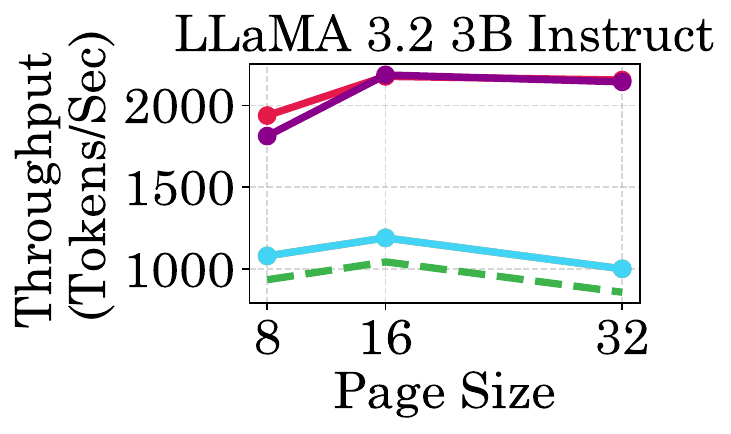}
        \caption{Throughput}
    \end{subfigure}
    \begin{subfigure}[b]{0.195\textwidth}
        \centering
        \includegraphics[width=\linewidth]{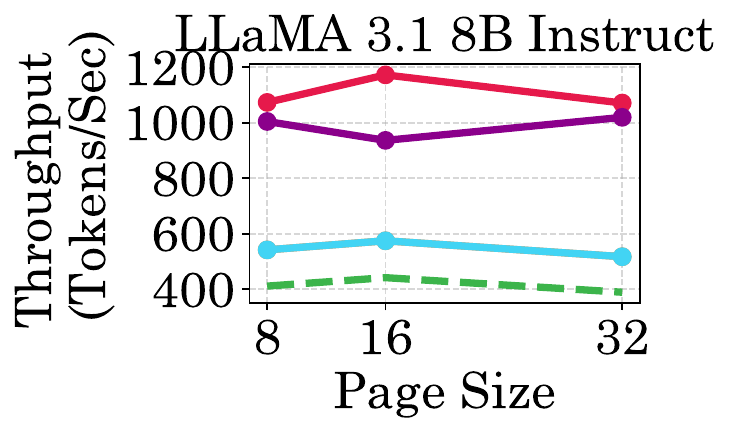}
        \caption{Throughput}
    \end{subfigure}
    \begin{subfigure}[b]{0.195\textwidth}
        \centering
        \includegraphics[width=\linewidth]{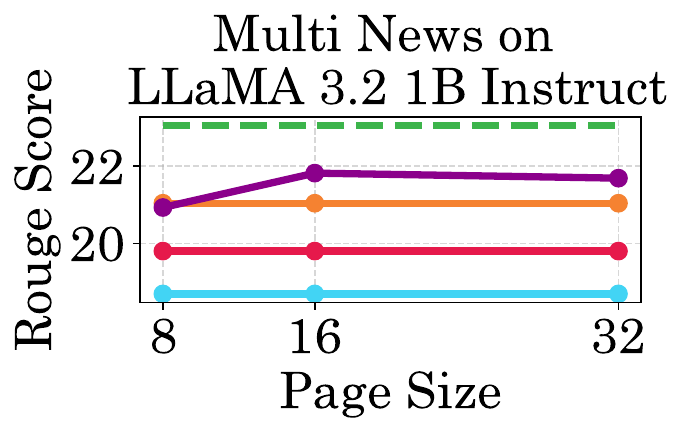}
        \caption{Rogue Score}
    \end{subfigure}
    \begin{subfigure}[b]{0.195\textwidth}
        \centering
        \includegraphics[width=\linewidth]{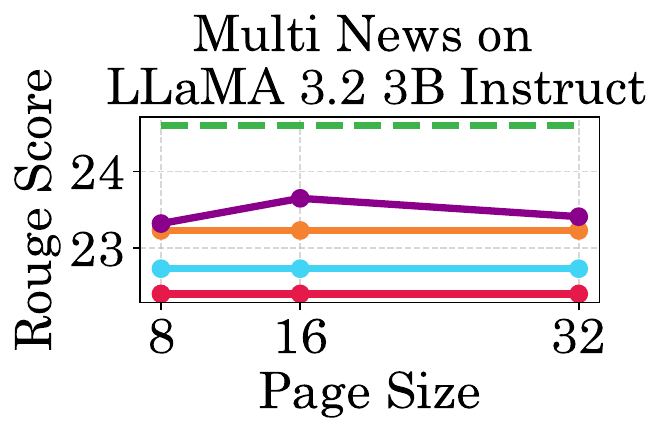}
        \caption{Rogue Score}
    \end{subfigure}
    \begin{subfigure}[b]{0.195\textwidth}
        \centering
        \includegraphics[width=\linewidth]{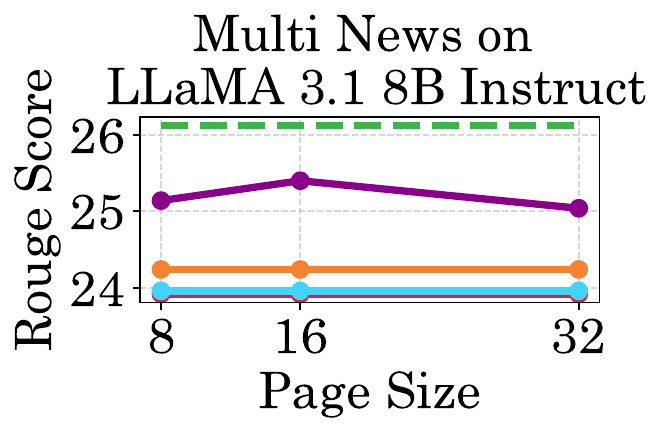}
        \caption{Rogue Score}
    \end{subfigure}
    \begin{subfigure}[b]{0.195\textwidth}
        \centering
        \includegraphics[width=\linewidth]{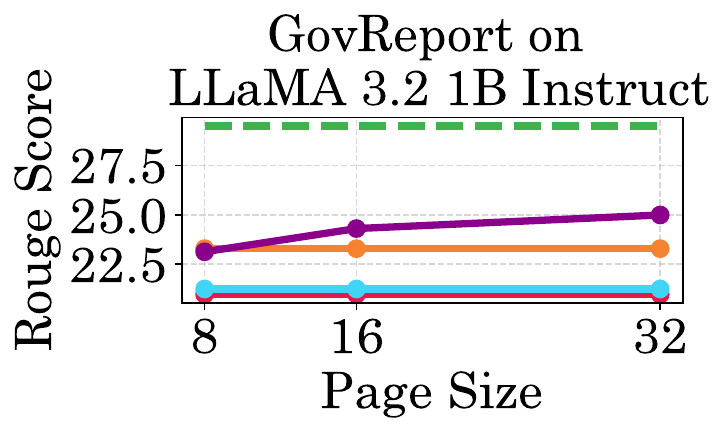}
        \caption{Rogue Score}
    \end{subfigure}
    \begin{subfigure}[b]{0.195\textwidth}
        \centering
        \includegraphics[width=\linewidth]{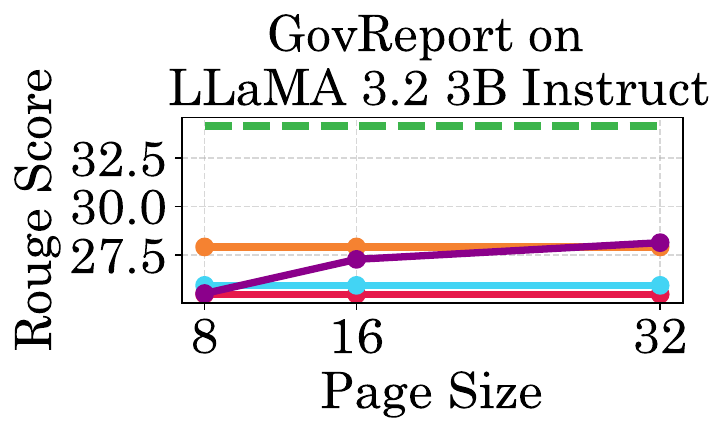}
        \caption{Rogue Score}
    \end{subfigure}
    \begin{subfigure}[b]{0.195\textwidth}
        \centering
        \includegraphics[width=\linewidth]{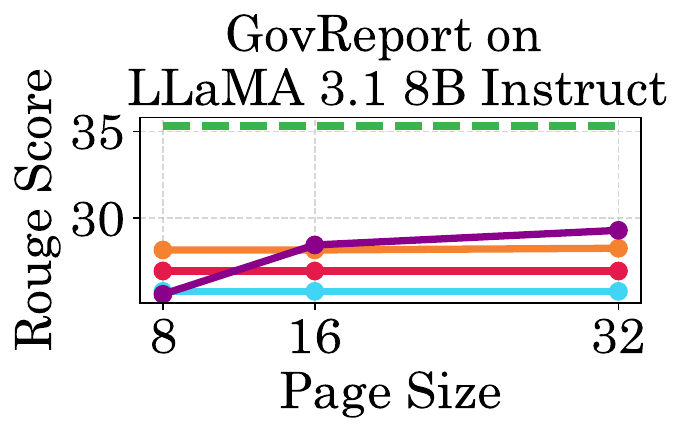}
        \caption{Rogue Score}
    \end{subfigure}
    \caption{Throughput (a–c) and ROUGE Score (d–i) across different LLaMA models (1B, 3B, 8B) and datasets (MultiNews, GovReport) for various KV cache compression methods and page sizes. PagedEviction consistently delivers high throughput—up to 3.1$\times$ over Full Cache while maintaining near-optimal accuracy with less than 3–5\% degradation. It outperforms Inverse Key L2-Norm and KeyDiff, which suffer from substantial accuracy drops.}

    \label{fig:page_sizes}
\end{figure*}

\subsection{vLLM Throughput and Latency}

Figures~\ref{fig:throughput_latency} (a), (b), and (c) illustrates the throughput results for the LLaMA-1B, LLaMA-3B, and LLaMA-8B models, respectively. PagedEviction consistently achieves better throughput than Inverse Key L2-Norm and KeyDiff while almost matching StreamingLLM across all cache budgets and models. For instance, at a cache budget of 1024, PagedEviction on LLaMA-1B model delivers $\sim$3020 tokens/sec, while StreamingLLM and Inverse Key L2-Norm/KeyDiff trail achieves only $\sim$2920 and $\sim$2170 tokens/sec, respectively, showing a 4.1\% and 39\% throughput improvement. PagedEviction outperforms Full Cache baseline (green dashed line at $\sim$2200 tokens/sec) by 37\% at cache budget 1024, while almost retaining the accuracy. Overall, these results validate that PagedEviction offers both higher efficiency and robustness across a wide range of memory budgets. StreamingLLM is straightforward to integrate into vLLM, as it simply requires zeroing out the last token and evicting the oldest block once it is full. However, it incurs overhead by evicting one token per decoding step and updating the KV cache block table at every step, which can be computationally expensive. In contrast, our proposed PagedEviction method performs eviction at fixed intervals, significantly reducing overhead. As a result, PagedEviction achieves comparable throughput to StreamingLLM despite some extra computation of calculating L2-Norms. On the other hand, unstructured eviction methods operate across the entire sequence and only evict a block once all its tokens have been individually evicted. This requires frequent checks across all blocks and prevents block-level eviction, offering little to no speedup during the decode phase. More visualizations of StreamingLLM and Unstructured Eviction (Inverse Key L2-Norm) and depicted in Appendix \ref{sec:appendix} (Figures \ref{fig:streamingLLM} and \ref{fig:random_pruning}). \looseness=-1 

Figure \ref{fig:throughput_latency} (d) depicts the time per output token across LLaMA models under cache budget of 1024. Our \textit{Paged Eviction} method consistently reduces latency compared to the Full Cache baseline, achieving reductions of approximately 12\%, 10\%, and 11\% for LLaMA-1B, 3B, and 8B models, respectively. Compared to StreamingLLM, Paged Eviction yields similar or slightly lower latency, indicating its efficiency despite less frequent evictions. The latency under Paged Eviction scales sublinearly with model size, highlighting its effectiveness in managing memory. \looseness=-1



\subsection{Ablation Study: Varying Page Sizes}

We conduct an ablation study by varying the page size and evaluating its impact on both accuracy and throughput. In Figure \ref{fig:page_sizes}, we compare throughput and model accuracy across two datasets for different page sizes under different KV cache compression methods. Across all settings, we consistently achieve a strong balance between throughput and accuracy. For instance, PagedEviction improves over Full Cache throughput by up to 3.1$\times$ and closely matches StreamingLLM, especially at page sizes 16 and 32. In terms of ROUGE score (d–i), PagedEviction yields less than 3–5\% degradation from Full Cache, outperforming other attention-free baselines. While KeyDiff and Inverse Key L2-Norm show significantly lower accuracy (20\% drop in (d)), we maintain robustness across both datasets and model scales, demonstrating its effectiveness on different page sizes.\looseness=-1

\section{Conclusion}

We propose PagedEviction, a structured block-wise KV cache eviction method for vLLM’s PagedAttention. Unlike existing methods that rely on attention scores or per-token comparisons, PagedEviction leverages a block-wise importance based on Key and Value states. This design preserves the structural integrity of memory blocks while minimizing overhead and enabling compatibility with CUDA attention kernels. Our method delivers high compression efficiency with minimal degradation in accuracy, achieving within 0.5-1.5 ROUGE points of full-cache performance at tight budgets (1024 tokens), while significantly improving throughput by up to 3.1$\times$ over Full Cache. Across LLaMA-1B, 3B, and 8B models, PagedEviction consistently yields 10–12\% lower latency, 15–20\% better accuracy than other methods like StreamingLLM, and robust results on long-context tasks in the LongBench benchmark. PagedEviction is easy to integrate, requires no CUDA attention kernel modification.

\section*{Acknowledgements}
This research used resources of the Argonne Leadership
Computing Facility, a U.S. Department of Energy (DOE)
Office of Science user facility at Argonne National Laboratory
and is based on research supported by the U.S. DOE Office
of Science-Advanced Scientific Computing Research Program,
under Contract No. DE-AC02-06CH11357

\bibliography{main.bib}

\appendix

\section{Appendix}\label{sec:appendix}


\subsection{StreamingLLM}

\begin{figure*}
    \centering
     \includegraphics[width=0.95\linewidth]{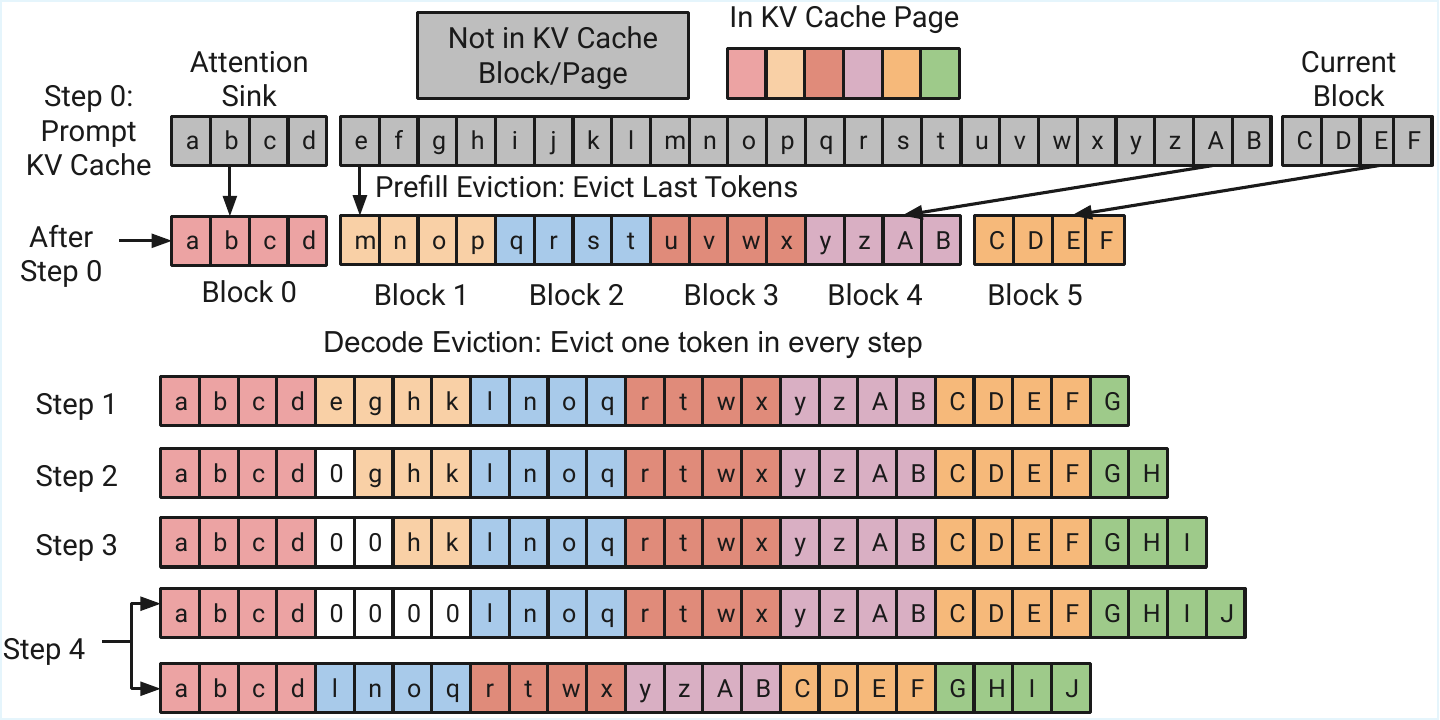}
        \caption{Illustration of the StreamingLLM KV cache eviction strategy. During prefill (Step 0), the last tokens are evicted to fit new blocks while preserving the first few tokens as attention sinks. During decoding, one token is evicted per step. The oldest block is completely evicted only when all tokens are evicted from the block. This approach ensures continuous streaming by maintaining a sliding window over the KV cache.}
        \vspace{-4mm}
        \label{fig:streamingLLM}
    \captionsetup{justification=centering}
\end{figure*}

Figure~\ref{fig:streamingLLM} illustrates the StreamingLLM eviction strategy. Initially, prompt tokens are processed, where only the earliest tokens (used as attention sinks) are retained, and the remaining prompt tokens are evicted based on recency. The KV cache is then organized into fixed-size blocks. In each decoding step, one token is evicted to make room for a new token, simulating a sliding window behavior. During decoding, one token is evicted per step. The oldest block is completely evicted only when all tokens are evicted from the block.

\subsection{Unstructured Eviction}

\begin{figure*}
    \centering
     \includegraphics[width=0.95\linewidth]{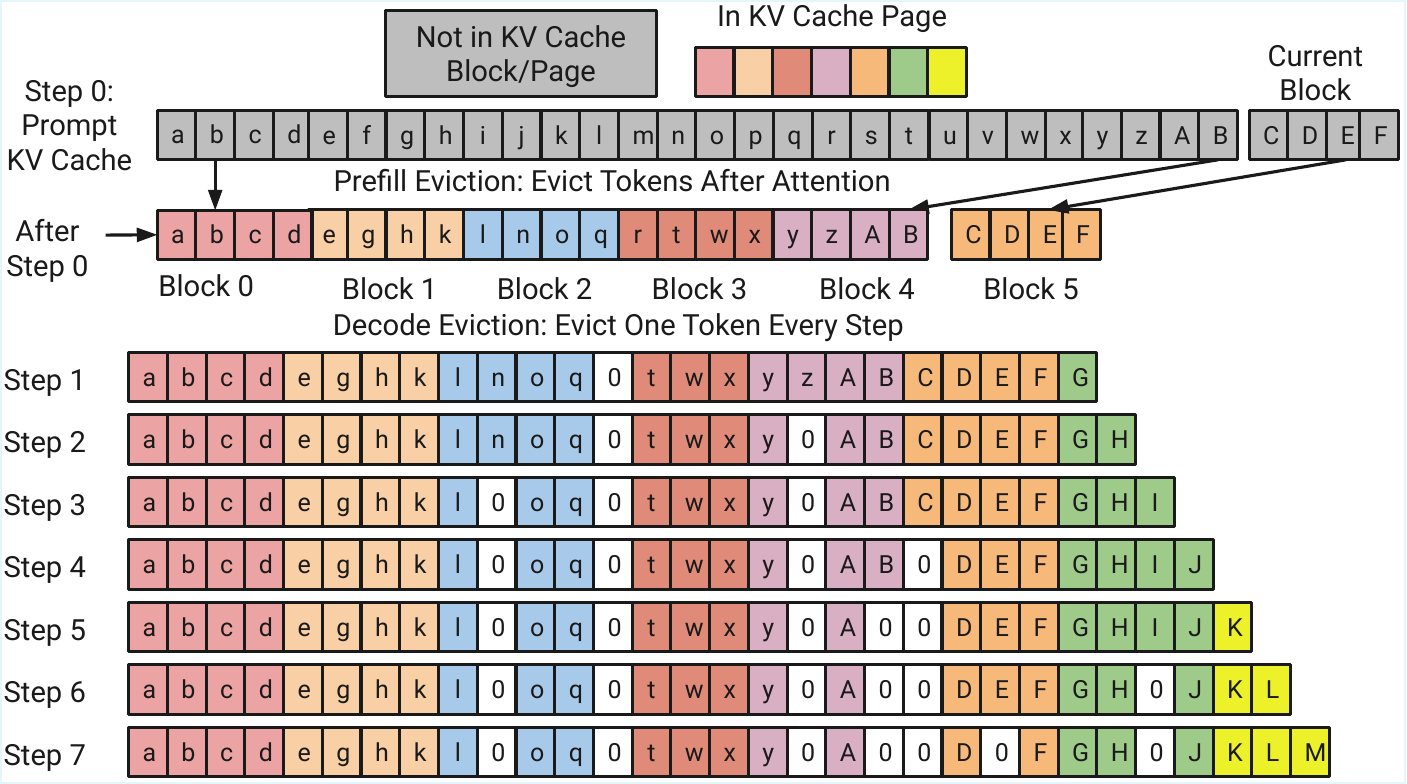}
        \caption{Illustration of Unstructured KV Cache Eviction using token-wise importance metrics such as Inverse Key L2-Norm or KeyDiff. During prefill (Step 0), low-importance tokens are evicted until we reach the cache budget. In the decode phase, one token is evicted per step based on its importance score, regardless of its position or block alignment. This leads to fragmented block occupancy, reducing the potential for full block eviction and resulting in inefficient memory usage.}
        \vspace{-4mm}
        \label{fig:random_pruning}
    \captionsetup{justification=centering}
\end{figure*}

Figure~\ref{fig:random_pruning} demonstrates the behavior of unstructured eviction methods such as Inverse Key L2-Norm or KeyDiff. Unlike structured strategies, unstructured eviction selects individual tokens for removal based on token-level importance scores, without regard for block structure. During prefill, tokens with lower importance are pruned. During decoding, the lowest-scoring token is removed in every step, which causes token-level fragmentation across blocks. This prevents entire blocks from being evicted efficiently, resulting in suboptimal memory utilization and requiring constant cache management.

\section{Related Work - KV Cache Optimization} \label{sec:related_work}

Here, we present a brief summary of previous works on KV cache optimization. For more details about KV Cache optimization and compression methods, please refer to surveys by Li et al. \cite{li2024survey} and Shi et al. \cite{shi2024keep}. 

\subsection{KV Cache Compression}


Various strategies have been developed to optimize KV cache memory by evicting less important tokens based on importance and relevance. H2O \cite{zhang2024h2o} employs accumulative normalized attention scores to retain high-impact tokens, known as Heavy Hitters, while ensuring recent tokens are preserved due to their strong correlations with current tokens. StreamingLLM \cite{xiao2023efficient} highlights the critical role of preserving key-value pairs from initial sequence tokens to maintain model performance. Keyformer \cite{adnan2024keyformer} addresses distortions in softmax probability distributions caused by token removal by introducing regularization techniques to smooth and approximate the original distribution. FastGen \cite{ge2023model} adopts a hybrid strategy, selecting token retention policies during prompt encoding (e.g., keeping special, punctuation, recent, or attention-weighted tokens) and applying these during decoding. SnapKV \cite{li2024snapkv} simplifies this by focusing solely on retrieving tokens based on importance scores, emphasizing that only a subset of prompt tokens is crucial for response generation. Scissorhands \cite{liu2024scissorhands} leverages the temporal significance of historically important tokens, preserving repetitive attention patterns through selective retention. \looseness=-1

\subsection{KV Cache Merge}

Several methods have been proposed to optimize KV cache storage by leveraging token similarity for merging. MiniCache \cite{liu2024minicache} identifies high angular similarity in KV caches of middle-to-deep layers and merges the Key and Value pairs of adjacent similar layers into a shared representation. D2O \cite{wan2024d2o} merges the Key or Value of evicted tokens with retained tokens based on cosine similarity. KVMerger \cite{wang2024model} clusters consecutive tokens with high cosine similarity to group contextually relevant tokens for merging. CaM \cite{zhangcam} uses attention scores to merge Keys or Values of multiple evicted tokens with retained tokens, producing a final merged representation. \looseness=-1

\subsection{Budget Allocation}


Many papers have explored hierarchical and adaptive strategies for KV cache management to optimize memory allocation and maintain model performance. PyramidInfer \cite{yang2024pyramidinfer} adopts a layer-wise approach, assigning more weight to recent tokens and using a decay ratio to reduce KV cache lengths in deeper layers, forming a pyramid structure. It also dynamically updates significant tokens during decoding based on attention values. PyramidKV \cite{zhang2024pyramidkv} employs a similar pyramid-shaped memory allocation strategy, allocating larger cache capacities to lower layers with uniform attention distributions and progressively reducing capacity in upper layers where attention is more concentrated on specific tokens. AdaKV \cite{feng2024ada} introduces head-specific memory allocation by leveraging distinct attention patterns across heads, optimizing cache distribution within a layer-wise budget using an L1 loss bound to preserve multi-head attention outputs.

\subsection{KV Cache Quantization}


There have been several quantization strategies proposed to optimize KV cache compression in LLMs while minimizing performance degradation. QAQ \cite{dong2024qaq} employs separate quantization for key and value caches, using an attention window to predict future attention scores and avoid over-compression of crucial tokens. GEAR \cite{kang2024gear} integrates uniform quantization, low-rank approximation for residuals, and sparse matrices to handle outlier errors. KVQuant \cite{hooper2024kvquant} addresses outliers by quantizing keys per channel before Rotary Positional Embedding (RoPE) and values per token, while retaining the first token in full precision to preserve performance. KIVI \cite{liu2024kivi} adopts mixed strategies by quantizing keys per channel and values per token, keeping recent KV pairs in full precision due to their critical role in token generation. MiKV \cite{yang2024no} uses mixed precision based on token importance, storing less significant KV pairs at a lower precision. ZipCache \cite{he2024zipcache} focuses on accurately computing token importance for efficient compression, while Atom \cite{zhao2024atom} applies fine-grained group quantization, handling outliers with higher precision and compressing normal channels to INT4 for maximum efficiency. 

\end{document}